\documentclass{llncs}

\usepackage{subcaption}
\usepackage{graphicx}
\usepackage{wrapfig}
\usepackage{tikz}
\usepackage[T1]{fontenc}
\usepackage{url}
\usepackage{multicol}
\usepackage{footmisc}
\usepackage{tcolorbox}
\usepackage{xcolor}

\usepackage[sectionbib,numbers,sort&compress]{natbib}
\usepackage{parskip}
\usepackage{caption}
\usepackage{subcaption}
\usepackage[hyperfootnotes=false]{hyperref}
\begin{document}

\title{Human-in-the-Loop and AI: Crowdsourcing Metadata Vocabulary for Materials Science}
%
%
\author{Jane Greenberg [0000-0001-7819-5360]\inst{1}
\and Scott McClellan [0000-0002-1524-8346]\inst{1}
\and Addy Ireland\inst{2}
\and Robert Sammarco\inst{1}
\and Colton Gerber [0000-0002-5471-0679]\inst{3}
\and Christopher B. Rauch\inst{1}
\and Mat Kelly [(0000-0002-0236-7389]\inst{1}
\and John Kunze [0000-0001-7604-8041]\inst{1}
\and Yuan An [0000-0001-9243-3572]\inst{1}
\and Eric Toberer [0000-0003-0826-2446]\inst{3}
}

\institute{Metadata Research Center, College of Computing and Informatics, Drexel University,
\email{\{jg3243,sm4522,rhs65,cr625,mrk335,jk3849,ya45\}@drexel.edu}
\and Penn State University, \email{addy.t.ireland@gmail.com}
\and Colorado School of Mines, \email{\{cgerber,etoberer\}@mines.edu}
}

\authorrunning{Greenberg et al.}
%
%
\maketitle              
\begin{abstract}
Metadata vocabularies are essential for advancing FAIR and FARR data principles, but their development constrained by limited human resources and inconsistent standardization practices. This paper introduces MatSci-YAMZ, a platform that integrates artificial intelligence (AI) and human-in-the-loop (HILT), including crowdsourcing, to support metadata vocabulary development. The paper reports on a proof-of-concept use case evaluating the AI-HILT model in materials science, a highly interdisciplinary domain Six (6) participants affiliated with the NSF Institute for Data-Driven Dynamical Design (ID4) engaged with the MatSci-YAMZ plaform over several weeks, contributing term definitions and providing examples to prompt the AI-definitions refinement. Nineteen (19) AI-generated definitions were successfully created, with iterative feedback loops demonstrating the feasibility of AI-HILT refinement. Findings confirm the feasibility AI-HILT model highlighting 1) a successful proof of concept, 2) alignment with FAIR and open-science principles, 3) a research protocol to guide future studies, and 4) the potential for scalability across domains. Overall, MatSci-YAMZ’s underlying model has the capacity to enhance semantic transparency and reduce time required for consensus building and metadata vocabulary development.

\end{abstract}

\section{Introduction}
The demand for actionable, high-quality metadata has drastically increased in connection with the rapid expansion of data-driven science and artificial intelligence (AI). Amid these changes, metadata vocabularies remain essential for core data management activities, including documentation, description, and overall life-cycle management \citep{wilkinson2016fair, SansoneSusanna-Assunta2019Faac}. Equally important is metadata's role in supporting the FAIR principles (Findability, Accessibility, Interoperability, and Reusability) across a wide range of research artifacts, such as data, code, algorithms, publications, AI prompts, and other supplemental materials \citep{wilkinson2016fair}. Recent attention to FARR (FAIR for machine learning, AI-Readiness, and Reproducibility) further highlights the central role that metadata plays in enabling transparent and trustworthy AI \citep{PraveenKumarReddyBandi_2025, Kirkpatrick2022, HuertaE.A.2023FfAA}. Collectively, these developments have transformed metadata into a respected, first-class data object, significantly diverging from the historical view of metadata as a purely practical or technical matter. Yet, despite this evolution, little has changed in how metadata vocabularies are developed, governed, and endorsed.

Metadata vocabulary development remains a complex and uneven activity shaped by resource limitations, epistemic expertise, and technical barriers. Additionally, metadata development in the sciences span a continuum from small academic laboratories, where vocabulary development is generally localized and informal, to large-scale industry and government settings, where the process is often formalized and time intensive. The range of practices, alongside human, financial, and infrastructure costs underscores the need for flexible, low barrier metadata vocabulary development platforms that ensure the necessary rigor of established standards. One promising approach is Yet Another Metadata Zoo (YAMZ), a crowdsourcing platform that enables researchers to propose, define, refine, and vote on terms collaboratively \citep{RauchChristopherB.2022FMAC}. YAMZ has successfully demonstrated the potential of community-driven approaches for materials science researchers; however, its reliance on human effort alone can be time-intensive \citep{GreenbergJane2023BCCf,deOliveiraIsabelM.2024Eaoa}. To address this limitation, we introduce Materials Science YAMZ (MatSci-YAMZ)---an extension of YAMZ that integrates AI and a human-in-the-loop (HILT) \citep{RetzlaffCarlOrge2024HRLA,ShneidermanBen2020HAIR} approach to balance scalability with expert validation and semantic accuracy \citep{Ireland2025}.

\section{Prescribed Languages and the Metadata Vocabulary Development Continuum}
Metadata vocabularies are semantic, ontological structures that offer a prescribed language supporting communication. Situating metadata vocabulary in scientific communication and a review of the Metadata Vocabulary Development Continuum provides context for the development of MatSci-YAMZ and the proof-of-concept use case presented in this paper.

\subsection{Language: A Fundamental of Scientific Communication}
Advances in scientific knowledge rely on communication in many forms—behavioral, oral, written, and increasingly computational \citep{alma991021824582304721,BazermanCharles1988Swk:}. While behavioral cues may assist collaboration, scientific progress remains primarily dependent on oral and written communication, both grounded in language. This shared linguistic foundation enables the development of discipline-specific terminology, which provides precision and clarity in scientific discourse \citep{alma991003744009704721,747902}. A shared semantics is also essential for the creation of metadata vocabularies that support documentation, representation, data-driven discovery, and AI. Among the most universally recognized examples are the taxonomic systems used for biological classification and the Periodic Table of Elements in chemistry—both function as prescribed languages that encode meaning for consistent interpretation. Every discipline has its own \textit{lingua franca}; some are more formally codified than others. Yet only a fraction of these vocabularies has been systematically encoded and made machine-readable as metadata systems, leaving substantial room for progress across the full spectrum of scientific research. 

Despite their paramount role, the development and maintenance of metadata vocabularies is often under-resourced and not always prioritized in scientific practice. One of the main reasons is that researchers often view metadata vocabulary development as trivial until they encounter documentation gaps, access issues, or interoperability barriers. Another factor is that the evolving nature of language requires continual attention to integrating new terms, while preserving historical terminology for reference. Interdisciplinary teams face additional complexity as they negotiate shared vocabularies across diverse conceptual frameworks. Moreover, developing, and sustaining metadata vocabularies requires human expertise to define and negotiate terms, plus continued human and financial investment, and infrastructure to support their technical implementation (this has been the focus of the FAIR 2.0 initiative \citep{VogtLars2024F2Et}). Adding to these challenges is the intensive time requirement necessary for the formalized standardization processes through organizations such as the National Information Standards Organization (NISO) and the International Organization for Standardization (ISO) \citep{GreenbergJane2023BCCf}. The formalized processes add rigor but require significant time and can extend over several years. These challenges vary across the scientific landscape and may interfere with scientific work.

\subsection{The Metadata Vocabulary Development Continuum}
\begin{figure}[h]
  \centering
  \includegraphics[width=.8\textwidth]{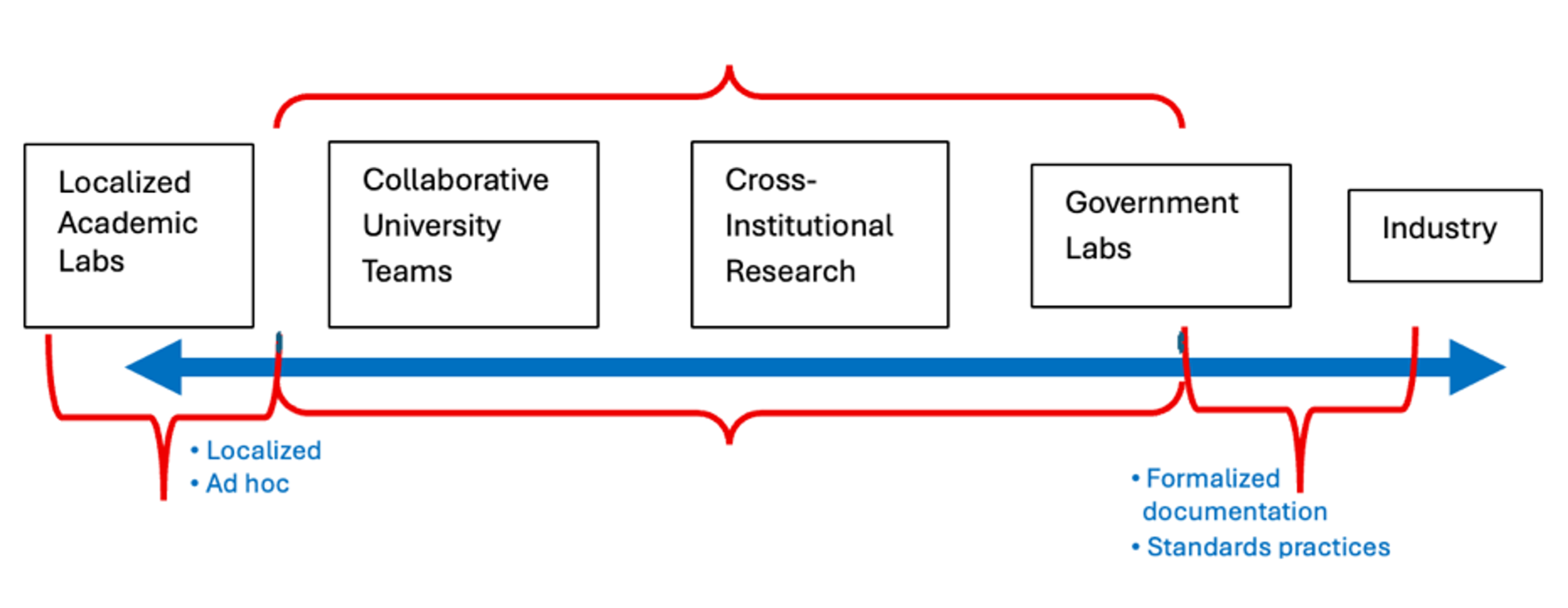}
  \caption{Metadata Vocabulary Development Continuum}
  \label{continuum}
\end{figure}

Metadata vocabulary development in science occurs along a continuum that reflects the diversity of research contexts, resources, and expertise (Figure \ref{continuum}). At one end are small-scale academic laboratories, where vocabulary work is often localized and ad hoc. Efforts are typically led by graduate students or postdoctoral researchers with limited training in metadata standards. The resulting vocabularies or data dictionaries tend to be project-specific, reflecting the immediate needs and constraints of the research group. Adoption of existing metadata vocabularies may occur when a researcher has prior experience or awareness of a system, but such cases are the exception rather than the rule.

At the other end of the continuum are large-scale industry and government laboratories. These organizations generally maintain more formalized metadata documentation practices, motivated by compliance requirements, regulatory expectations, or competitive advantage. Large labs can often afford off-the-shelf software to implement specific metadata. Such institutions may participate in formal standards processes coordinated by the ISO, NISO, or discipline-specific consortia like the International Union of Crystallography's Crystallographic Information Framework (CIF). However, these processes are often resource-intensive, time-consuming, and lack transparency to the broader research community.
\vspace{-1.5em}
This uneven landscape highlights persistent challenges in vocabulary development, including staff turnover, inconsistent documentation, and the human effort required to achieve consensus. These challenges are especially pronounced in interdisciplinary fields such as materials science, where research innovation continually introduces new terms and concepts. Furthermore, terminology is sometimes repurposed to address field-specific concerns \citep{alma991003744009704721}. Even cases where definitions vary slightly for a single term among adjacent fields can impede communication to varying degrees, ranging from mild confusion to misunderstanding. Lastly, neologisms may emerge, that are niche, unclear, or unstable. Addressing these complexities requires low-barrier approaches that balance the rigor necessary for shared metadata vocabularies with the desired flexibility \citep{McClellanScott2024CSAo}.

Crowdsourcing offers one such model by leveraging collaborative, networked communication. The YAMZ platform exemplifies this approach, enabling participants to propose, discuss, and vote on preferred term definitions—helping to democratize consensus-building for metadata vocabularies. While YAMZ has demonstrated community engagement and scalability potential, its reliance on sustained human participation remains a key limitation. Integrating AI into the YAMZ workflow introduces new opportunities for efficiency while maintaining semantic accuracy and expert oversight. These considerations motivate the present study, which explores how collaborative HILT methods and AI-assisted vocabulary generation may strengthen and accelerate metadata vocabulary development.

\section{Goals and Objectives}
\label{sec:objectives}

This study responds to the challenges reviewed above, an accelerated interest in interdisciplinary research, and the emerging opportunity to leverage AI. The overall goal is to explore how humans and AI can collaboratively develop a mutual, agreed-upon understanding of terminology for metadata vocabularies, as an essential component of scientific research.

 The specific objectives of the research reported on in this paper are as follows:
\begin{enumerate}
\item Extend and adapt the YAMZ platform (including the original YAMZ-MatSci extension \citep{Sammarco2025}) into MatSci-YAMZ, an environment where AI can collaborate with researchers to refine term definitions.
\item Demonstrate a hybrid approach for vocabulary development that combines crowdsourcing, AI, and HILT oversight.
\item Evaluate materials science as a domain testbed for exploring vocabulary challenges in interdisciplinary contexts.
\item Develop a research protocol to guide further analysis and reproducibility.
\end{enumerate}

Together, these objectives frame a proof-of-concept case study that investigates an AI–HILT-supported model for metadata vocabulary development.

\section{Research Design: Methods and Procedures}
To investigate the the overall goal and objectives outlined above we conducted a proof-of-concept case study \citep{Elliott_2021}. A proof-of-concept study allows researchers to explore how new processes or systems operate in practice before broader implementation. The case study method, widely applied in information science and sociotechnical research, provides a systematic means of examining technological interactions and interconnected activities in a real-world context. The case tested the feasibility of combining crowdsourcing, AI, and HILT methods to support metadata vocabulary development. Materials science served as the application domain, selected for its interdisciplinary nature and persistent challenges in harmonizing terminology across subfields such as physics, chemistry, and engineering.

The MatSci-YAMZ \citep{Ireland2025} core model replicates the general YAMZ application and incorporates features of the YAMZ-MatSci prototype \citep{Sammarco2025}. MatSci-YAMZ is coded in Python and uses PostgreSQL for data management. Users add terms, examples, and definitions, and can up-vote and down-vote the definitions. The AI feature requires that the user enter the term and provide an example. The example helps to disambiguate context and initiates the generation of the AI definition. MatSci-YAMZ also implements a provenance tracking feature for changes made to all definitions.

\subsection{Procedures}
\begin{figure}[h]
  \centering
  \includegraphics[width=.9\textwidth]{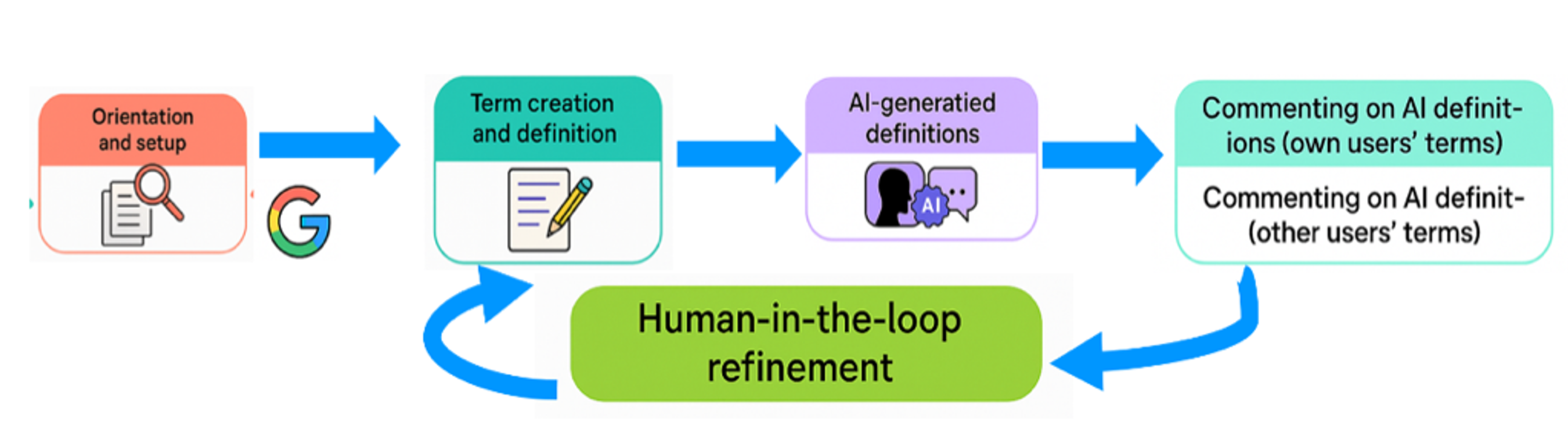}
  \caption{Research Procedures Workflow}
  \label{procedures}
\end{figure}

Participants engaged with the MatSci-YAMZ platform (https://matsci.yamz.net) through a structured workflow designed to explore AI-HILT interaction in vocabulary generation and refinement. Six (6) participants were recruited for this study. The participants included materials science researchers and those with computational experience working with materials scientists. All participants were  affiliated with the Institute for Data-driven Dynamical Design (ID4), an NSF Harnessing the Data Revolution Institute. The procedures included the following steps:
\begin{enumerate}
    \item \textbf{Orientation and setup:} Participants first reviewed tutorial slides explaining the platform's core features and workflow. A brief video recording was also made available to supplement the orientation.
    \begin{itemize}
        \item \textit{Setup activity:} Participants created a MatSci-YAMZ account using their Google credentials to enable personalized interaction with the system.
    \end{itemize}
    \item \textbf{Term entry and definition:} Each participant was instructed to add two (2) terms relevant to their research domain. For each term, they provided an initial definition and an illustrative example.
    \item \textbf{AI-generated definitions:} The system then used the Gemma3 model to automatically generate AI-based definitions for each term entered by the user, based on the definitions and examples provided by the user.
    \item \textbf{Commenting on AI generated definitions:}
    \begin{itemize}
        \item \textit{Commenting AI definitions (own terms):} Participants reviewed the AI-generated definitions associated with their own terms, providing comments and evaluative feedback regarding clarity,
        \item \textit{Commenting on AI definitions (other users' terms):} Participants were also asked to comment on AI-generated definitions created for other users’ terms, fostering peer review and cross-disciplinary reflection.
    \end{itemize}
\end{enumerate}
These procedures (visually captured in Figure \ref{procedures}) set up the capacity for a HILT refinement loop, whereby AI-generated definitions could be iteratively improved through human review and AI revisions.

\section{Results}
\begin{figure}[h]
  \centering
  \includegraphics[width=.7\textwidth]{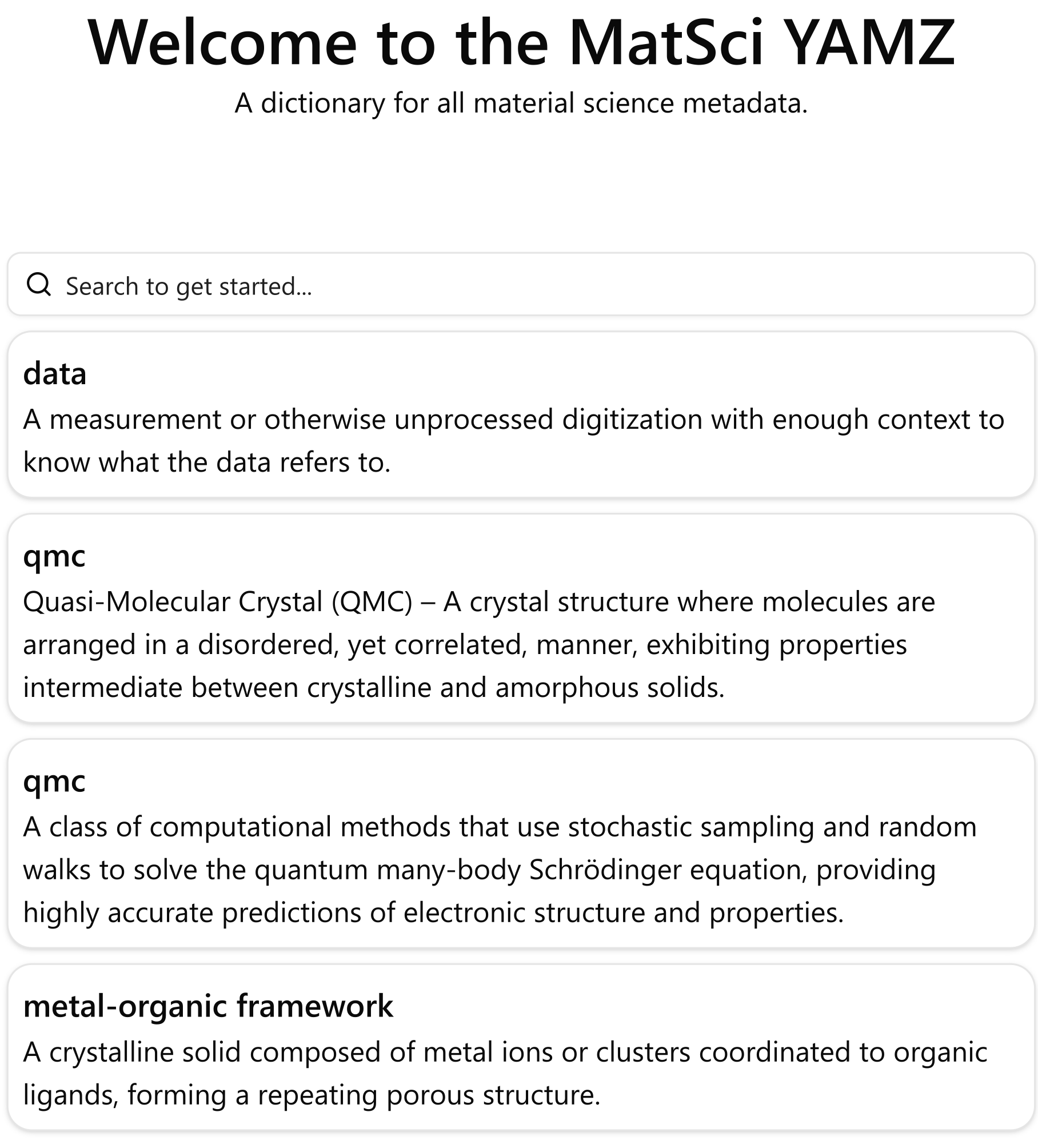}
  \caption{MatSci YAMZ Welcome Page}
  \label{welcome}
\end{figure}

\begin{figure}[h]
  \centering
  \includegraphics[width=.7\textwidth]{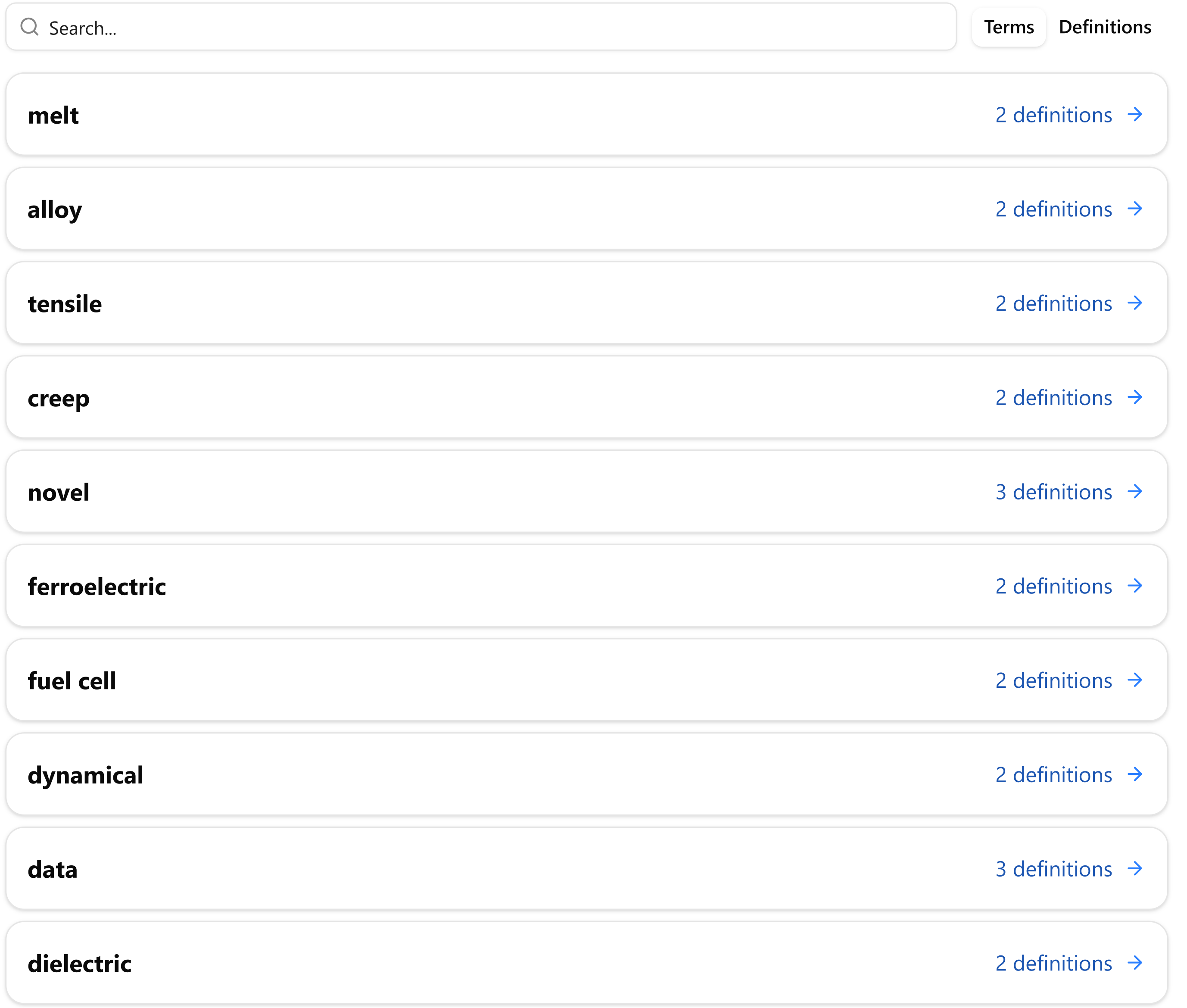}
  \caption{MatSci YAMZ top directory}
  \label{top}
\end{figure}

Over the testing period of several weeks (late July to mid-September 2025, with specific tasks required during a 2-week period toward the end of August 2025), the six participants collectively entered 20 human-generated terms accompanied by definitions and illustrative examples. Of the initial 20 terms and examples, MatSci-YAMZ successfully produced 19 AI-generated definitions, with one instance where a term had two human-generated definitions.  
Figure \ref{welcome} shows the MatSci-YAMZ Welcome Page, which supports an initial search, where users see an initial definition; and Figure \ref{top} shows the recently updated MatSci-YAMZ top directory of terms, where user browse, add terms, click, and explore further to tag the terms or add terms and definitions.
\subsection{MatSci-YAMZ}

\begin{figure}[ht]
  \centering
  \includegraphics[width=.8\textwidth]{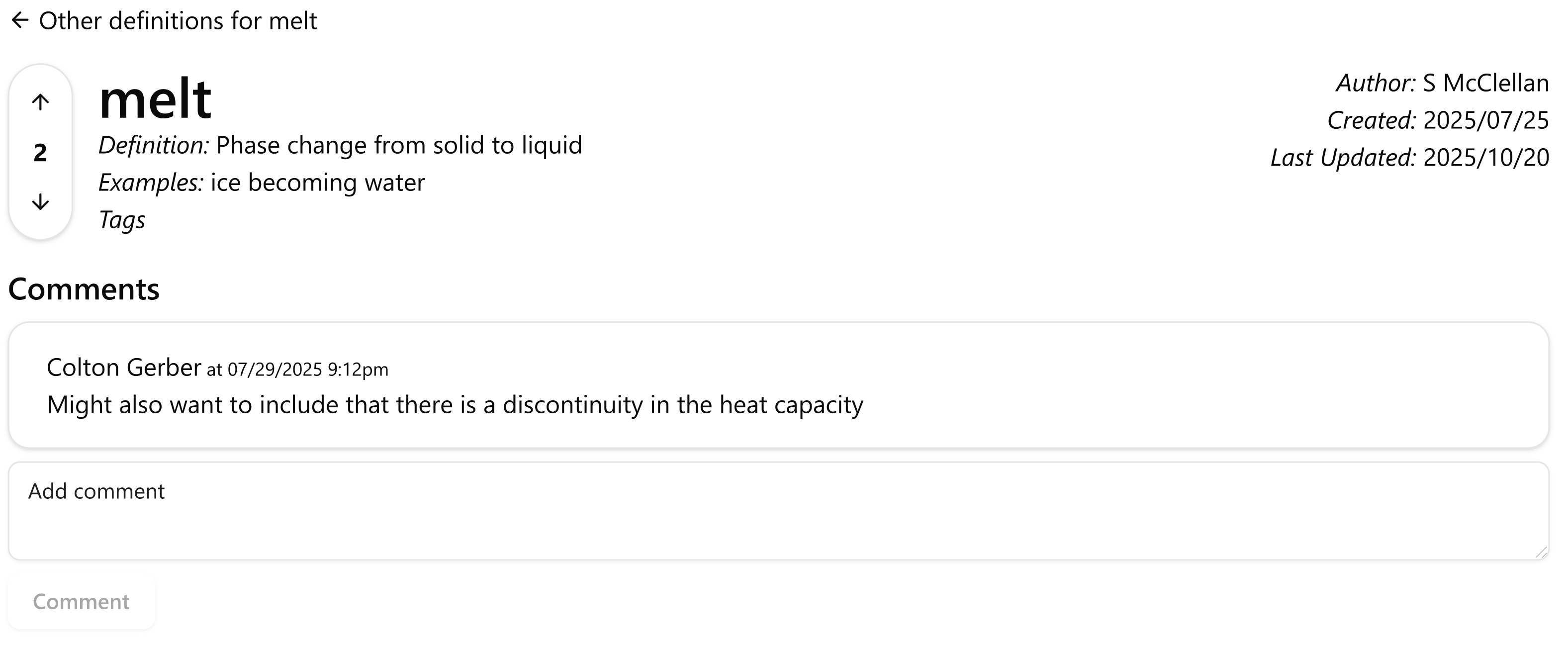}
  \caption{Initial seed entry and scientist comment for the term ``melt''}
  \label{melt}
\end{figure}

\begin{figure}[h]
\centering
\includegraphics[width=.8\textwidth]{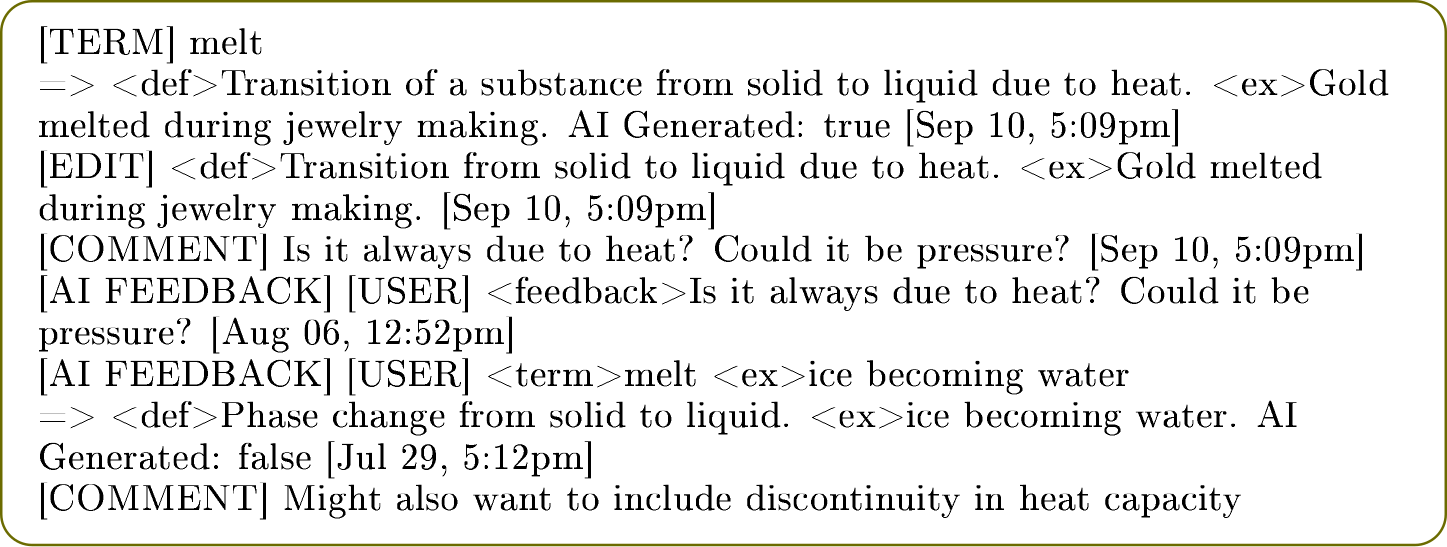}
\caption{Provenance view for term ``melt''}
\label{fig:provenance}
\end{figure}

\begin{figure}[t]
\centering
\includegraphics[width=.8\textwidth]{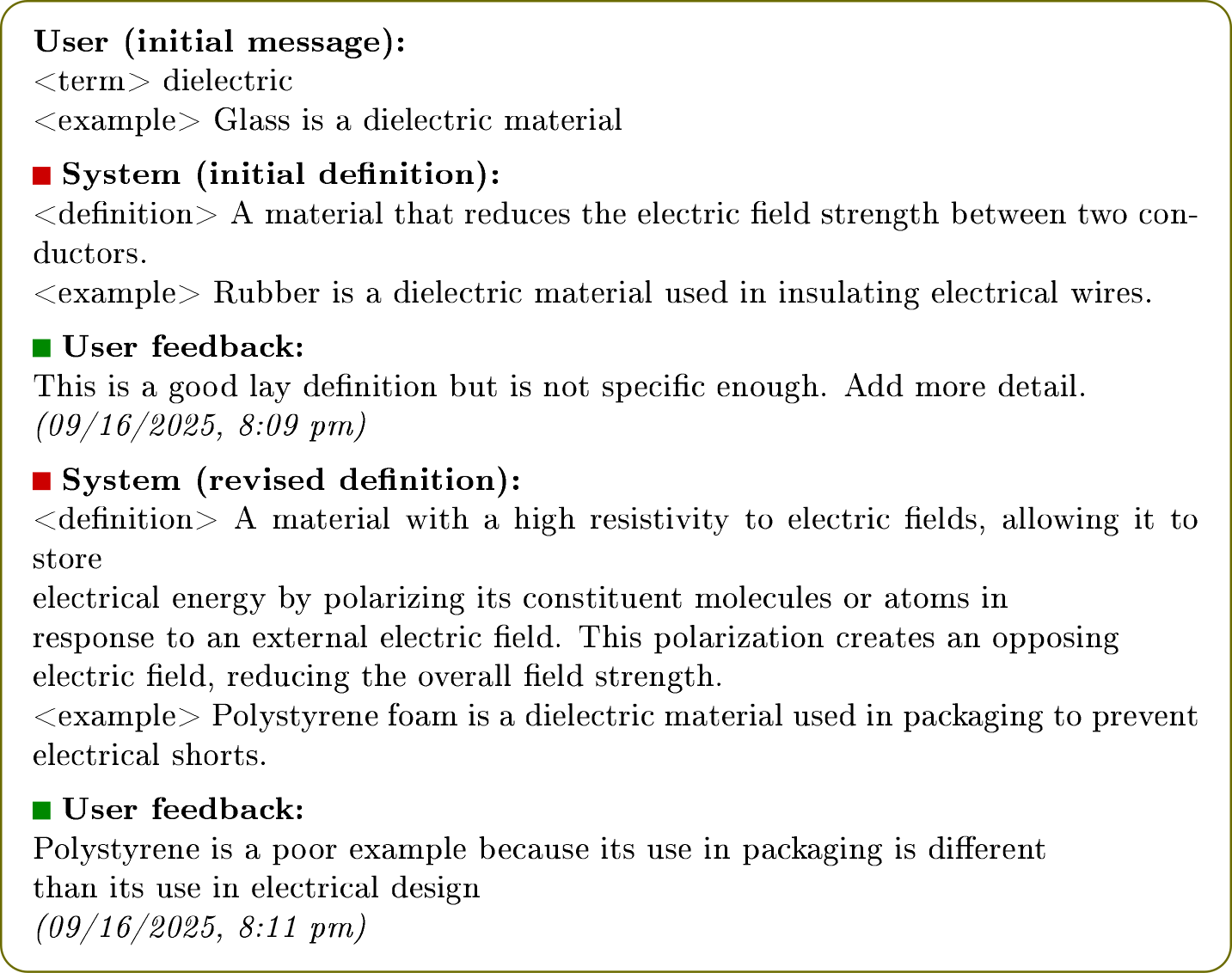}
\caption{AI and User Negotiation}
\label{fig:ai_negotiation}
\end{figure}

 Prior to the launch of the study, MatSci-YAMZ was seeded with several terms, drawing from previous definitions contributed in previous YAMZ testing \citep{GreenbergJane2023BCCf}. Figure \ref{melt} shows an entry for ``melt'' that was added July 25, 2025, to support the study, with a comment made on July 29, 2025. This figure also shows the last update was relatively recently, October 20, 2025, due to some additional testing and interaction.
 A more involved sequence tracking the provenance of the term development, is captured in Figure \ref{fig:provenance}. The sequence is read bottom to top, by date to denote each action. The log starts on July 29, 2025, and continues through September 10, 2025. The ``regular text'' represents human interaction, while the ''\textbf{bold text}'' captures AI activity.
 A more user-friendly sequence log documenting the user experience is presented in Figure \ref{fig:ai_negotiation}, which reads top to bottom. In this example, the participant, identified as ``user,'' initially submits the term ``dielectric'' along with an example. The example initiates the AI sequence to generate a definition. An AI activity is designated by the ``red'' square, and a user activity is designated by the ``green'' square. The back-and-forth captures the AI-HILT negotiation to establish a term's definition. In this example, the user is not fully in agreement with the AI. Whereas, the results were promising with AI helping to refine and specify initial human-generated definitions other instances.  

 The current iteration of MatSci-YAMZ permits a single AI-generated definition per term along with continued AI-HILT interaction and definition revision. As reported above, the open test period ran for several weeks with specific tasks required during a 2-week period in late August 2025.  Despite the testing timeframe, the provenance logs show continued participation in MatSci-YAMZ after the designated test period, as both initial participants and new users continue to explore this platform. The continued user activity is an indication of the platform's usability and potential value of MatSci-YAMZ's for metadata vocabulary development.

\section{Discussion}
The MatSci-YAMZ proof-of-concept case study demonstrates that a hybrid AI-HILT approach can effectively augment existing crowdsourced models of metadata vocabulary development. The results suggest that this model may support the establishment of metadata vocabularies while reducing some of the significant overhead that researchers face across the Metadata Vocabulary Development Continuum. More broadly, these findings provide insight into the objectives that guided this work, which are addressed here:
\begin{itemize}
    \item \textit{Extending and adapting YAMZ into MatSci-YAMZ:} The case study confirmed that YAMZ (including the YAMZ-MatSci prototype) was extended and adapted to an AI-HILT platform supporting collaborative metadata vocabulary development. MatSci-YAMZ allows users to define terms, contribute examples, and observe and negotiate with AI to refine a term definition. Additionally, the example-based prompting proved instrumental in contextualizing terms and generating meaningful AI definitions.
    \item \textit{Demonstrating a hybrid model combining crowdsourcing, AI and HILT oversight:} The AI-HILT workflow facilitated iterative improvement of several term definitions and revealed some disagreements with the AI. The platform's provenance-tracking feature further supported transparency and oversight by recording all edits, comments, and system responses for later analysis. The up-vote and down-vote further supports the crowdsourcing, collaborative feature, and offers a mechanism for consensus building.
    \item \textit{Evaluating materials science as a testbed for interdisciplinary vocabulary challenges:} Materials science served as an effective testbed for AI-assisted vocabulary generation due to its inherently interdisciplinary nature. Although the participant group was small, each person had a distinct area of expertise. Materials science depends on clear communication across a number of disciplines (e.g., physics, chemistry, and engineering, math, and other interconnected domains), and productive collaboration requires a shared understanding of the nuanced terminology within and across these various topic areas.
    \item \textit{Developing a research protocol to guide future reproducibility:} A key contribution of this study is the establishment of a protocol for future testing AI-HILT workflows supporting crowdsourced vocabulary development. The structured procedure, including the orientation, term entry, AI generation, and multi-user commentary, can be replicated or extended for subsequent studies in materials science, in lab, a designated research community, an institutional context, or an open environment. Future developments of AI-HILT YAMZ in other disciplines may also benefit from exploring this framework. 
\end{itemize}



The results also allow for additional observations on how term definitions evolve and are refined over time. Provenance data offered insights into how human and AI agents can collaboratively delineate the meaning of a term. The edit histories and timestamps revealed how definitions changed in response to both human feedback and AI suggestions. While this test was exploratory and chiefly focused on the mechanics of MatSci-YAMZ as a proof of concept, the AI, prompted by an example, generated a few acceptable definitions that were voted up, although in other instances the dialog and feedback loop was not completed (e.g. \ref{procedures} above). Collectively, these observations highlight an opportunity for improving prompt engineering to assist and accelerate the development of metadata vocabularies.

As with any study, there are limitations due to practical research constraints that warrant acknowledgment. This project was conducted as a proof-of-concept case study designed to test the feasibility of integrating AI and human-in-the-loop (HILT) methods to support metadata vocabulary development. The study was intentionally exploratory, given the recent launch. Additionally, the limited number of participants and the short timeframe likely influenced the results. Despite these factors, the results provided valuable insight into how an AI-HILT model can aid metadata vocabulary development and engage community members.

\section{Conclusions}

This study introduced and tested MatSci-YAMZ, a crowdsourcing, AI-HILT platform designed to support metadata vocabulary development. The proof-of-concept use case demonstrated that MatSci-YAMZ's underlying model has the capacity to leverage AI and the ``wisdom of the crowds'' \citep{alma991022064629004721}, through crowdsourcing, to support vocabulary development. Four key contributions emerge from this work:
\begin{itemize}
    \item \textit{Proof of concept:} MatSci-YAMZ confirms the viability of combining AI-HILT and crowdsourcing to support and potentially accelerate the time required for metadata vocabulary development.
    \item \textit{Alignment with FAIR and open-science principles:} MatSci-YAMZ (and the general YAMZ model) contribute to advancing the FAIR principles and open science. Provenance tracking and crowdsourced features further promote accountability and reproducibility across evolving terminologies.
    \item \textit{A replicable research protocol:} The research protocol that framed this study can guide future testing of AI-HILT models and workflows, including those with crowdsourcing, to support metadata vocabulary development.
    \item \textit{Scalability and extensibility:} The combination of prompt engineering with examples, and AI-definition refinement, with the HILT feedback loop and crowdsourcing, has potential cross-domain expansion and broader community-driven vocabulary initiatives.

\end{itemize}

Future work will extend testing with a larger participant cohort in materials science, integrate refined provenance analytics and additional AI models, study the voting of term definitions, and implement a persistent identifier system for term reuse. Sustained user engagement following the proof-of-concept use case indicates continued interest and potential for platform growth. Together, these contributions indicate that MatSci-YAMZ has potential for supporting metadata vocabulary development, an essential component of today's dynamic, evolving landscape of scientific communication.

\section*{Acknowledgments}
The authors acknowledge partial support by the National Science Foundation (NSF) under grant NSF-HDR-OAC 2118201 Institute for Data Driven Dynamical Design.

\bibliography{MatSci_YAMZ}
\bibliographystyle{splncs04}

\end{document}